\documentclass{article}
\usepackage{spconf,amsmath,graphicx}

\usepackage{diagbox}

\usepackage[usestackEOL]{stackengine}
\usepackage{amsmath}
\usepackage{graphicx}
\usepackage{multirow}
\usepackage{multicol}
\usepackage{amssymb}% http://ctan.org/pkg/amssymb
\usepackage{pifont}% http://ctan.org/pkg/pifont

\usepackage{hyperref}
\usepackage{cleveref}[capitalise]

\usepackage{caption}

\DeclareMathOperator*{\argmax}{argmax}

\newcommand{\cmark}{\ding{51}}
\newcommand{\xmark}{\ding{55}}

\title{Instant One-Shot Word-Learning for Context-Specific\\ Neural Sequence-to-Sequence Speech Recognition}
\name{Christian Huber$^1$, Juan Hussain$^1$, Sebastian St\"uker$^1$ and Alexander Waibel$^{1,2}$}
\address{
  $^1$Interactive Systems Lab, Karlsruhe Institute of Technology, Karlsruhe, Germany\\
  $^2$Carnegie Mellon University, Pittsburgh PA, USA\\~\\
  firstname.lastname@kit.edu, alexander.waibel@cmu.edu}
  
\begin{document}
%\ninept
%
\maketitle

\begin{abstract}
Neural sequence-to-sequence systems deliver state-of-the-art performance for automatic speech recognition (ASR). When using appropriate modeling units, e.g., byte-pair encoded characters, these systems are in principal open vocabulary systems. In practice, however, they often fail to recognize words not seen during training, e.g., named entities, numbers or technical terms. To alleviate this problem we supplement an end-to-end ASR system with a word/phrase memory and a mechanism to access this memory to recognize the words and phrases correctly. After the training of the ASR system, and when it has already been deployed, a relevant word can be added or subtracted instantly without the need for further training. In this paper we demonstrate that through this mechanism our system is able to recognize more than 85\% of newly added words that it previously failed to recognize compared to a strong baseline.
\end{abstract}


\section{Introduction}
Up until recently automatic speech recognition (ASR) systems were implemented as Bayes classifiers in order to search for the word sequence $\hat{W}$, among all possible word sequences $W$, with the highest posterior probability given a sequence of feature vectors $X$ which is the result of pre-processing the acoustic signal to be recognized:% $\hat{W}=\argmax_W P(W|X)=\argmax_W P(X|W)P(W)$
% \begin{equation*}
%     \hat{W}=\argmax_W P(W|X)=\argmax_W P(X|W)P(W)
% \end{equation*}
\begin{align*}
    \hat{W}&=\argmax_W P(W|X)\\
    &=\argmax_W P(X|W)P(W)
\end{align*}
In the context of ASR $P(X|W)$ is called the acoustic model, $P(W)$ the language model. The space of allowed word sequences to search among was usually defined by a list of words, the vocabulary, of which permissible word sequences could be composed. Words that were not in the vocabulary could not be recognized. In turn this means that by adding words to the vocabulary and appropriate probabilities to the language model, previously unknown words could be easily added to the ASR system manually or even automatically.

In contrast, for neural sequence-to-sequence trained end-to-end ASR systems, this is not possible anymore. While in principal end-to-end systems are open-vocabulary systems, when using appropriate modeling units, such as byte-pair encoded (BPE) characters, in practice, words not seen during training are often not reliably recognized. This is especially true for named entities. The reasons for that are that, a) the end-to-end network implicitly learns language model knowledge when being trained on transcribed speech data, and b) especially named entities often have a grapheme-to-phoneme relation that deviates from the general pronunciation rules of the language, as learnt implicitly by the networks of the end-to-end system.

In order to solve this problem, in this paper we extend an end-to-end ASR system by a memory for words and phrases. We further introduce a mechanism that enables the system to recognize the words stored in this memory without further training. This enables us to add new words and short phrases (only text, no audio) to the recognition system, long after the training of the ASR system has finished, and when it has already been deployed, without the need for further training.

This is achieved by, a) a memory-attention layer which
predicts the availability and location of relevant information in the memory,
%contains i) a gate layer which outputs if there is useful information for the prediction of the next token in the memory and ii) an attention layer which output where in the memory this information is
and b) a memory-entry-attention layer which extracts the information of a memory entry.

%we extend a sequence-to-sequence model by a memory component and a mechanism to access it to learn new words. The model learns during training if there is useful information in the memory (via a gate layer), where this information is (via memory-attention layer) and how to use it (via a memory-entry-attention layer). During inference a new word can be added to the memory. Only the text of the new word is required but no audio. This process does not involve any training. After that the model automatically attends to the new memory entry if needed and the new word can be recognized correctly.

\begin{figure*}[!h]
  \centering
  \includegraphics[page=5,width=\textwidth,trim=0.9cm 3.0cm 8.4cm 1.5cm,clip]{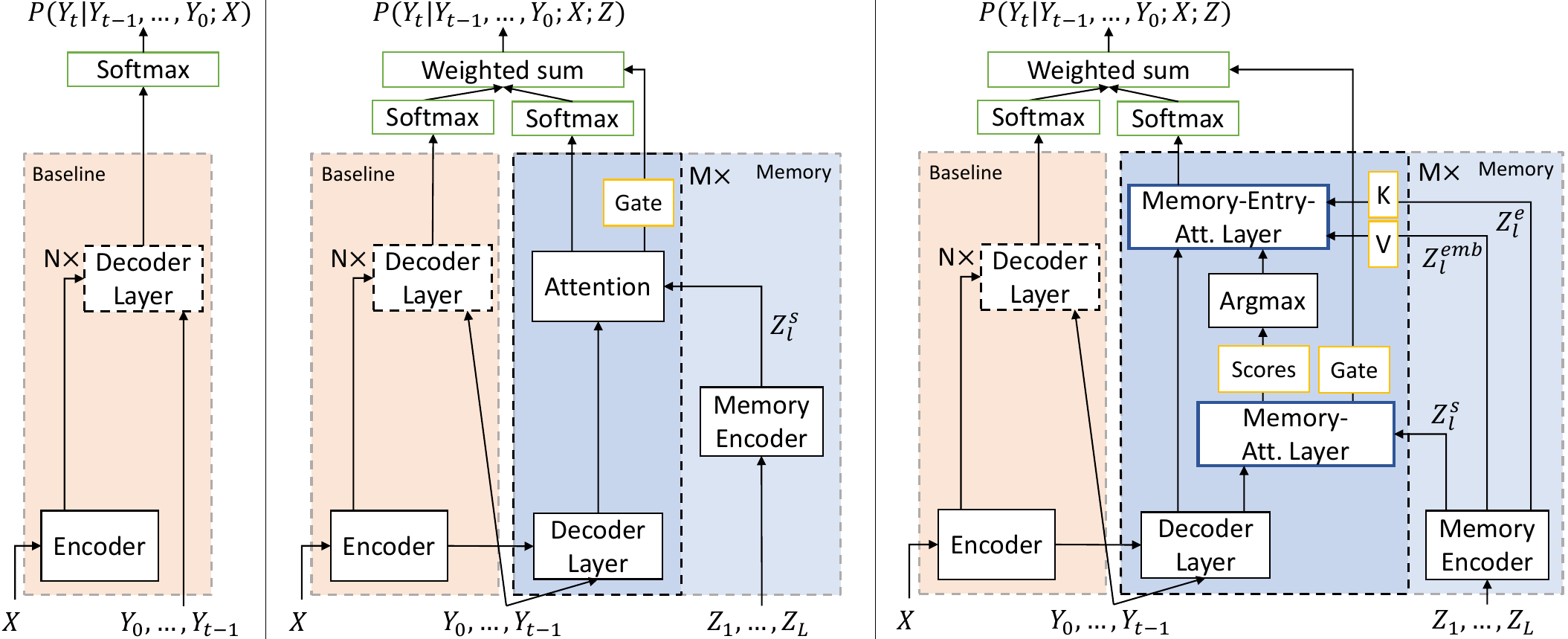}
  \caption{Left: Baseline model, Middle: MEM model similar to \cite{pundak2018deep}, Right: 2MEM model. Dark blue: The two main components that are added compared to the baseline.}
  \label{fig:model}
  %\vskip -5pt
\end{figure*}

\section{Related Work}
The problem of new or rare words in speech recognition with sequence-to-sequence models is an active field of research.
%The closest to our work is \cite{pundak2018deep}, a spelling model which outputs graphemes supplied by dynamically biased context of rare words.
The closest to our work is \cite{pundak2018deep}. In their work, they train a LSTM-based ASR model based on \cite{chan2016listen} which outputs graphemes. The model is supplied by dynamically biased context of rare words.
\cite{bruguier2019phoebe} injects additional phonetic information on top of that. \cite{coucheiro2019attention} predicted a vector presenting the context of a word with a low confidence (misrecognized) and searched for the nearest vectors in a big corpus to find a substitution. In \cite{feng2017memory} a neural machine translation is augmented by a memory of word pairs to translate unseen or sparse pairs in the training data, but only for single words to be translated.

In our work, we build a memory for new words and phrases. The first modeling of a memory was done with Recurrent Neural Networks (RNNs) and Long-Short Term Memory (LSTMs), which capture long-term structure within the sequence in a latent state (local memory). In praxis, many applications require addressable memory (global memory) for the neural network in a kind of Neural Turing Machines \cite{graves2014neural}.
For instance, Memory Networks \cite{weston2014memory} and End-To-End Memory Networks \cite{sukhbaatar2015end} are used to store answers for question-answering applications. Similarly, Pointer Networks \cite{vinyals2015pointer} circumvent the out-of-vocabulary (OOV) problem by copying words from the input sequence that are not contained in the output vocabulary. The network does not possess any memory. A well-known example of a Pointer Network is the summarization with Pointer-Generator Networks \cite{see2017get}.
\cite{santoro2016one} build a memory-augmented neural network for meta-learning which saves sample-label pairs and predicts upon the similarity of the input and the samples encoded in the memory. The most frequent memory samples are injected into the weights as long-term memory.
\cite{chao2020automatic} aim to automatically spell-correct ASR output text using a pointer network to copy out-of-vocabulary (OOV) words directly. They incorporate phonetic information as well, however they do not mention any kind of attending to audio as in our work. 

Other related works address the problem of adaptation to a new domain or enhance the model by rescoring the output with a language model via two pass decoding. These works does not aim to solve the problem of OOV words directly. \cite{raju2019scalable} incorporate a multi corpora language model for second pass rescoring, while \cite{gandhe2020audio} and \cite{sainath2019two} rescore with a second model by attending to the audio or as in \cite{hu2020deliberation}, where the second model attends to both the audio and the output using a deliberation model.

\section{Method}
\label{sec:method}

\subsection{Baselines}
\label{sec:baseline}

As baseline we use a transformer-based sequence-to-sequence model \cite{vaswani2017attention,pham2019very}. It consists of 24 encoder layers and 8 decoder layers with a model dimension of 1024 and feed-forward dimension of 2048 (see \cref{fig:model}, left). The inputs $X$ to the encoder are mel-filterbank coefficients with 40 features. Before the encoder layers a two-layer convolutional neural network with 32 channels and time stride two is used to downsample the input spectrogram \cite{nguyen2019onfly}. The decoder has as input the BPE tokens of the already decoded sequence $Y_0,\ldots,Y_{t-1}$ and predicts a probability distribution $P(Y_t|Y_0,\ldots,Y_{t-1};X)$ over the BPE vocabulary. The BPE consists of 4000 tokens.

For comparison we trained a memory-enhanced model (MEM) similar to \cite{pundak2018deep} (see \cref{fig:model}, middle). However we replaced the LSTMs they used with a transformer model. Furthermore we run the baseline decoder together with the memory decoder and combine the output at the end (see \cref{sec:mem-dec}). This has the advantage that we are able to change the baseline after the training.

Our proposed model\footnote{A demo can be found under \url{https://huberchristian.eu}} (see \cref{fig:model}, right) changes the memory-decoder compared to the MEM model such that the output is computed in a two-step fashion. First, the availability and location of relevant information is predicted, and second, the information of a memory entry is extracted.
We call this the two-step memory-enhanced model (2MEM). Both the MEM model and the 2MEM model predict a probability distribution $P(Y_t|Y_0,\ldots,Y_{t-1};X;Z)$ over the BPE vocabulary, where $Z=(Z_1,\ldots,Z_L)$ are additional words/phrases we want the model to recognize.

To demonstrate that our model is able to be used with any baseline model we also evaluate it using a LSTM-based baseline \cite{nguyen2019onfly}. This baseline consists of 6 encoder layers and 2 decoder layers with a model dimension of 1536.

\subsection{Memory Encoder}
\label{sec:mem-enc}

Our memory can contain (similar to \cite{pundak2018deep}) words or phrases in each memory entry. We represent each memory entry as BPE and refer to them as $Z_1,\ldots,Z_L$, where \linebreak
$Z_l\in\{1,\ldots,|V_{BPE}|\}^{e_l}$, $l\in\{1,\ldots,L\}$. The term $e_l$ denotes the number of tokens in the memory entry $Z_l$.
%The indices $Z_l$ are in a set containing $|BPE|+3$ elements because we add to the BPE tokens the padding, begin and end of sequence tokens.

The memory encoder starts with an embedding layer which output is denoted by $Z_l^{emb}\in\mathbb{R}^{e_l\times d_{model}}$, where $d_{model}$ is the dimension of the model.
Then a transformer encoder is applied. It consists of eight layers, if not mentioned otherwise. The output of the encoder is denoted by $Z_l^e\in\mathbb{R}^{e_l\times d_{model}}$.

After that we compute the mean of each encoded memory entry $Z_l^e$ over the BPE tokens of the entry. This leaves us with one vector per memory entry, which can be interpreted as summary vector of the memory entry, denoted by \linebreak
$Z_l^s\in\mathbb{R}^{d_{model}}$.
We add to these vectors $Z_l^s$ a learned dummy vector $Z_0^s$, which is later used to determine when there is no relevant information in the memory.

\subsection{Memory Decoders}
\label{sec:mem-dec}

Both the decoders in the MEM model and the 2MEM model consist of $M$ blocks. In our experiments we use $M=4$ if not mentioned otherwise.
Each block starts with a decoder layer similar to the baseline model. Then a memory-attention layer is applied. It extracts the availability and location of relevant information of the memory.
%The query input $q$ and the key input $k$ are a linear transformation of the decoder layer output and $Z_l^s$, respectively.
This is done by calculating similarity scores, where for the query input $Q$ and the key input $K$ the decoder layer output and $Z_l^s$ are used, respectively.
\begin{align*}
    \hat{Q} &= W^QQ,\qquad&\text{ shape }T\times d_{model}\\
    \hat{K} &= W^KK,\qquad&\text{ shape }(L+1)\times d_{model}\\
    scores &= \hat{Q} \hat{K}^T,\qquad&\text{ shape }T\times(L+1)\\
    gate &= scores[:,0],\qquad&\text{ shape }T
\end{align*}
where $T$ is the number of tokens in the target sequence, $W^Q$ and $W^K$ are weight matrices and $scores[:,0]$ are the similarity scores corresponding to the learned dummy vector $Z_0^s$.

The main difference between the MEM model and the 2MEM model is the way in which the information is extracted of the memory. For the MEM model all the information of a memory entry has to be encoded in the vector $Z_l^s$. For the 2MEM model the vector $Z_l^s$ has to only contain the information if a memory entry is currently relevant. If thats the case the memory-entry-attention layer can extract the information.

In the MEM model the score output of the memory-attention layer is processed by a softmax layer and the result is multiplied with a linear transformation of values $V$ generated by $Z_l^s$.
\begin{align*}
    \hat{V} &= W^VV,\qquad&\text{ shape }(L+1)\times d_{model}\\
    output &= sm(scores)\hat{V},\qquad&\text{ shape }T\times d_{model}
\end{align*}
where $W^V$ is a weight matrix, $sm$ is the softmax function, which is computed w.r.t. the last dimension. The memory-attention layer together with this transformation is equivalent to an attention layer similar to the encoder-decoder attention. The difference is that the gate output is used in the further processing.

In the 2MEM model the memory-entry-attention layer is used to extract relevant information of a specific memory entry. This is done by using an attention mechanism similar to the encoder-decoder attention, however not the vectors $Z_l^s$ are used to generate the values as in the MEM model.
For the query again the decoder layer output is used and for the keys and values, we use $Z_l^e$ and $Z_l^{emb}$, respectively, where $l$ is the argmax of the scores of the memory-attention layer for a given query.

Let ${l=\argmax(scores_t)}$ be the index of the largest score for the query $Q_t$ at time step $t$. If $l$ is not zero, i.e. the index does not correspond to the learned dummy vector, the following is computed:
\begin{align*}
    \hat{Q} &= \Bar{W}^QQ_t,\qquad&\text{ shape }d_{model}\\
    \hat{K} &= \Bar{W}^KZ_l^e,\qquad&\text{ shape }e_l\times d_{model}\\
    \hat{V} &= \Bar{W}^VZ_l^{emb},\qquad&\text{ shape }e_l\times d_{model}\\
    output &= sm(\hat{Q}\hat{K}^T)\hat{V},\qquad&\text{ shape }T\times d_{model}
\end{align*}
where $\Bar{W}^Q$, $\Bar{W}^K$ and $\Bar{W}^V$ are weight matrices.
If the maximum score corresponds to the learned dummy vector (see \cref{sec:mem-enc}), no attention is calculated. A residual connection is used around the attention layer in the MEM model and the memory-entry-attention layer in the 2MEM model.

For both the MEM model and the 2MEM model, the outputs of the baseline decoder and the memory decoder are combined at the end by a weighted sum. The weighting is calculated as a linear transformation with one output neuron of the stacked gate outputs of all memory attention layers followed by a sigmoid layer.

\section{Training}
During training we freeze the original baseline components and train only the memory related components.

\subsection{Memory Content during Training}
\label{sec:mc}
The content of the memory during training is sampled for each batch dependent on the target labels of the batch. We randomly select up to three words of a target label and use these as a memory entry. In total we sample 200 memory entries for each batch.

\subsection{Loss Function}
\label{sec:lf}

The loss function $L$ consists of two cross entropy parts. The first one classifies the next token $Y_t$ of the sequence and the second one classifies the most important memory entry:
\begin{align*}
    L&=\frac{1}{T}\sum_{t=1}^TCE(f_\theta(Y_{t-1},\ldots,Y_0;X;Z),Y_t)\\
    &+\frac{\lambda}{MT}\sum_{m=1}^M\sum_{t=1}^TCE(scores_t^m,label^{mem}_t),
\end{align*}
where $f_\theta(Y_{t-1},\ldots,Y_0;X;Z)$ is the output distribution of the decoder given the already decoded sequence $Y_{t-1},\ldots,Y_0$, the audiofeatures $X$ and the memory/context $Z=(Z_l)_l$.
Furthermore, $scores_t^m\in\mathbb{R}^{L+1}$ are the score output of the memory attention layer in the $m$-th memory decoder block and $label_t^{mem}\in\{0,\ldots,L\}$ is the memory label. This memory label is constructed when sampling the memory content (see \cref{sec:mc}). It contains the index zero of the learned dummy vector if a token is not chosen for the memory and otherwise the index of the corresponding memory entry.
We use $\lambda=1$ and label smoothing of $0.1$.

We experienced that the MEM model learns to only focus on the baseline decoder output. Therefore, we force the model to use the memory when advantageous.
This is done by permuting the output of the baseline decoder and the memory decoder dependent on $label^{mem}$ with a certain probability before the weighted sum is applied.
This forces the model to use the memory decoder output if there is relevant information in the memory and the baseline decoder output otherwise.
In particular, the probabilities of the label index and a random other index of the baseline decoder output are interchanged where $label^{mem}$ is non-zero and the same is done for the memory decoder output where $label^{mem}$ is zero.
From the permuted probabilities no gradient is propagated backwards to not confuse the model.

\subsection{Data}

For training and evaluation of our models, we used Mozilla Common Voice v6.1 \cite{ardila2019common}, Europarl \cite{koehn2005europarl}, How2 \cite{sanabria2018how2}, Librispeech \cite{panayotov2015librispeech}, MuST-C v1 \cite{mustc19}, MuST-C v2 \cite{cattoni2021must} and Tedlium v3 \cite{hernandez2018ted} datasets.
The data split is presented in the following \cref{table:asr-english-data}.
\begin{table}[htbp]
\begin{tabular}{|l|r|r|} \hline
 Corpus & Utterances & Speech data [h]\\ \hline\hline
 \multicolumn{3}{|l|}{\textbf{A: Training Data}} \\
 \hline\hline
Mozilla Common Voice  & 1225k & 1667 \\ %1224864
Europarl & 33k & 85\\ %32628
How2 & 217k & 356\\ %217358
Librispeech & 281k & 963\\ %281241
MuST-C v1 & 230k & 407\\ %229702\\
MuST-C v2 & 251k & 482\\ %250941
Tedlium & 268k & 482\\ %268263
\hline\hline
 \multicolumn{3}{|l|}{\textbf{B: Test Data}} \\
 \hline\hline
Tedlium & 1155 & 2.6\\
%Librispeech & 2620 & 5.4\\
New-words testset & 239 & 0.5\\
\hline
\end{tabular}
\caption{\label{table:asr-english-data} Summary of the data-sets used. }
\end{table}

Furthermore we created a new-words testset containing 239 utterances each containing a word, e.g. named entities, we assumed the model could do wrong because it has not seem these words during training. We use this new-words testset to evaluate the performance of the models to learn these new words when they are provided as additional context through the memory. 

%TODO: examples?

\section{Results}
\label{sec:results}

\subsection{Evaluation}
For the evaluation we report an accuracy on new-words testset with empty and full memory as well as the WER on the ted testset also with empty and full memory. As memory entries we use all the (239) new words of the new-words testset. The accuracy is thereby calculated such that the output of the model is counted as correct if the new word is present in the transcript.

The results can be seen in \cref{table:results}.
We obtained that the MEM model learned to only focus on the baseline decoder. A reinitialization of the parameters in the weighted sum during training did not help. This changes when permuting the output probabilities of the decoder (see \cref{sec:lf}) with probability 0.5. The accuracy on the new-words testset increases from 44.8\% to 79.5\%.

Surprisingly, the 2MEM model works well on the new-words testset without permuted output probabilites (88.7\%), however there is also a performance gain when permuting the output probabilities of the 2MEM model (90.4\%).

On the ted testset the performance drops slightly for all models with good performance on the new-words testset. This happens since the models sometimes uses a word in the memory when the target word starts similarly.

%TODO: Why MEM, 2MEM better? 2MEM better suited for longer memory entries

\begin{table}[th]
  \centering
  \resizebox{1.00\columnwidth}{!}{%
    \begin{tabular}{|l|c|c|c|c|} \hline
    \multicolumn{1}{|r|}{Metric} & \multicolumn{2}{c|}{\Centerstack{Acc.$\uparrow$ (\%)\\ new words testset}} & \multicolumn{2}{c|}{\Centerstack{WER$\downarrow$ (\%)\\ ted testset}}\\
    \hline
    \multicolumn{1}{|c|}{\backslashbox{Model}{Full memory}} & \xmark & \cmark & \xmark & \cmark\\
    \hline\hline
    Transformer baseline & \multicolumn{2}{c|}{45.2} & \multicolumn{2}{c|}{5.0}\\
    \hline
    MEM model & 44.8 & 44.8 & 4.8 & 4.8\\
    \multicolumn{1}{|r|}{+ permute 0.5} & 44.4 & 80.8 & 5.0 & 5.3\\
    \hline
    2MEM model & 42.3 & 88.7 & 4.9 & 5.2\\
    \multicolumn{1}{|r|}{+ permute 0.5} & 45.2 & \textbf{90.4} & 5.0 & 5.2\\
    \hline
    \end{tabular}
  }
\caption{\label{table:results} Summary of the results. %For some models results from two epochs are shown. 
  Accuracy on the new-words test set and WER on the ted test set. }
\end{table}

In \cref{example} an example of correcting the sentence "[we will] hopefully have covid nineteen behind [us soon]" with the model 2MEM with permutation probability 0.5 can be seen. The word covid nineteen was not seen in the training data.

\begin{table}[th]
  \centering
    a) Decoding with empty memory
    \resizebox{\columnwidth}{!}{
    \begin{tabular}{|c|c|c|c|c|c|c|c|c|c|} \hline
    hope & fully & have & co & ve & t & nin & et & een & behind\\
    \hline
    0.90 & 0.88 & 0.89 & 0.83 & 0.89 & 0.93 & 0.93 & 0.88 & 0.88 & 0.89\\
    \hline
    0 & 0 & 0 & 0 & 0 & 0 & 0 & 0 & 0 & 0\\
    \hline
    \end{tabular}
    }
    \vskip 10pt
    b) Decoding with word "covid" as $Z_1$
    \resizebox{\columnwidth}{!}{
    \begin{tabular}{|c|c|c|c|c|c|c|c|c|c|} \hline
    hope & fully & have & co & v & id & nin & et & een & behind\\
    \hline
    0.90 & 0.88 & 0.87 & \textbf{0.03} & \textbf{0.03} & \textbf{0.03} & 0.92 & 0.88 & 0.88 & 0.89\\
    \hline
    0 & 0 & 0 & \textbf{1} & \textbf{1} & \textbf{1} & 0 & 0 & 0 & 0\\
    \hline
    \end{tabular}
    }
    \caption{\label{table:example} Example for correcting "covid nineteen". Rows: Predicted tokens, gate (weighting between baseline decoder and memory decoder) and memory location output (argmax of the scores; all layers have the same argmax in this example). }
    \label{example}
\end{table}

\subsection{Ablation}
\label{sec:ablation}

For the ablation we start with the best performing model, the 2MEM model with permutation probability $0.5$, and explore different changes. The results can be seen in \cref{table:results2}.

\begin{table}[th]
  \centering
  \resizebox{1.00\columnwidth}{!}{%
    \begin{tabular}{|l|c|c|c|c|} \hline
    \multicolumn{1}{|r|}{Metric} & \multicolumn{2}{c|}{\Centerstack{Acc.$\uparrow$ (\%)\\ new words testset}} & \multicolumn{2}{c|}{\Centerstack{WER$\downarrow$ (\%)\\ ted testset}}\\
    \hline
    \multicolumn{1}{|c|}{\backslashbox{Model}{Full memory}} & \xmark & \cmark & \xmark & \cmark\\
    \hline\hline
    LSTM baseline & \multicolumn{2}{c|}{47.3} & \multicolumn{2}{c|}{3.9}\\
    \hline
    \Centerstack{2MEM model, permute 0.5} & 45.2 & 90.4 & 5.0 & 5.2\\
    \multicolumn{1}{|r|}{+ encode values} & 44.8 & 90.4 & 4.9 & 5.2\\
    \multicolumn{1}{|r|}{+ residual connection} & 44.8 & 88.3 & 5.1 & 5.3\\
    \multicolumn{1}{|r|}{+ LSTM baseline} & 46.9 & \textbf{92.1} & 3.9 & \textbf{4.2}\\
    \multicolumn{1}{|r|}{\Centerstack{one memory\\ encoder layer}} & 43.1 & 84.5 & 5.3 & 6.0\\
    \multicolumn{1}{|r|}{\Centerstack{one memory\\ decoder layer}} & 42.7 & 87.9 & 5.2 & 7.1\\
    \hline
    \end{tabular}
  }
\caption{\label{table:results2} Summary of the ablation results. %For some models results from two epochs are shown. 
  Accuracy (in \%) on the new-words test set, WER on the Ted test set. }
\end{table}

We tried to use $Z_l^e$ instead of $Z_l^{emb}$ (see sections \ref{sec:mem-enc} and \ref{sec:mem-dec}) as value input for the memory-entry-attention layer. We refer to this as encode values and obtained that using these values delivers almost the same performance.
Furthermore, we tried to add residual connections to the scores of the memory-attention layer. The performance on the new-words testset as well as on the ted testset decreased.
As already mentioned in \cref{sec:baseline} our model can be used with any baseline. We substituted our transformer-based baseline with a LSTM-based one which has a better performance on the ted testset. We obtain that this performance increase transfers to the new-words testset.
We also trained models with only one memory encoder or one memory decoder layer. We see that this also works on the new-words testset, however the performance on the ted testset degrades to 6.0\% and 7.1\%, respectively, when the new words are in the memory. It seems that the memory decoder layers are more critical than the memory encoder layers.

Overall the 2MEM model with LSTM-based baseline performed the best with 92.1\% accuracy on our new-words testset. This model was able to correct more than 85\% of the errors via the memory component.

%Decoding schemes

\section{Conclusion and Future Work}
\label{sec:conclusion}

In this paper we demonstrated that it is possible to enhance a sequence-to-sequence ASR model with a memory component that allows to add new words to the recognition system, e.g., named entities, during deployment without the need for further training. In this way more than 85\% of the previously misrecognized words not seen during training, were recognized correctly. Our 2MEM model outperforms the MEM model (which is similar to the model in \cite{pundak2018deep}) significantly.

One drawback of the proposed method is that one only provides text for the memory. Therefore, if the pronunciation of a new word deviates from the implicitly learned pronunciation rules, it is unlikely to be found by the memory-attention layer. We therefore plan to extend our model by populating the memory with an audio sample of the new word in addition to text. 
Furthermore, a subword regularization \cite{kudo2018subword}, where a word can be encoded into multiple BPE candidates, could improve the performance of the system.

\section{Acknowledgements}
The projects on which this paper is based were funded by the Federal Ministry of Education and Research (BMBF) of Germany under the numbers 01IS18040A and 01EF1803B. %The authors are responsible for the content of this publication.

\bibliographystyle{IEEEbib}
\bibliography{refs}

\end{document}